\documentclass{article}

\usepackage[square,sort,comma,numbers]{natbib}

\usepackage[preprint]{neurips_data_2022}

\usepackage[utf8]{inputenc} % allow utf-8 input
\usepackage[T1]{fontenc}    % use 8-bit T1 fonts
\usepackage{hyperref}       % hyperlinks
\usepackage{url}            % simple URL typesetting
\usepackage{booktabs}       % professional-quality tables
\usepackage{amsfonts}       % blackboard math symbols
\usepackage{nicefrac}       % compact symbols for 1/2, etc.
\usepackage{microtype}      % microtypography
\usepackage{xcolor}         % colors

\usepackage{colortbl}
\definecolor{Gray}{gray}{0.9}
\definecolor{Red}{rgb}{0.95, 0.8, 0.8}
\definecolor{Green}{rgb}{0.8, 0.95, 0.8}
\definecolor{Blue}{rgb}{0.8, 0.8, 0.95}

\usepackage{epsfig}
\usepackage{graphicx}
\usepackage{amsmath}
\usepackage{amssymb}
\usepackage{xfrac} 

\usepackage{caption}

\usepackage{multirow}
\usepackage{tablefootnote}

\usepackage{wrapfig}

\newcolumntype{L}[1]{>{\raggedright\arraybackslash}m{#1}}
\newcolumntype{C}[1]{>{\centering\arraybackslash}p{#1}}

\usepackage{pifont}% http://ctan.org/pkg/pifont
\newcommand{\cmark}{\ding{51}}%
\newcommand{\xmark}{\ding{55}}%

\title{Vision meets mmWave Radar: 3D Object Perception Benchmark for Autonomous Driving}

\author{%
 Yizhou Wang\textsuperscript{1,*}, Jen-Hao Cheng\textsuperscript{1,*}, Jui-Te Huang\textsuperscript{2}, Sheng-Yao Kuan\textsuperscript{3},\\
 \textbf{Qiqian Fu\textsuperscript{4}, Chiming Ni\textsuperscript{4}, Shengyu Hao\textsuperscript{4}, Gaoang Wang\textsuperscript{4},} \\
 \textbf{Guanbin Xing\textsuperscript{1}, Hui Liu\textsuperscript{1}, Jenq-Neng Hwang\textsuperscript{1}}\\
 \textsuperscript{1}{University of Washington},
 \textsuperscript{2}{Carnegie Mellon University},\\
 \textsuperscript{3}{National Yang Ming Chiao Tung University}, 
 \textsuperscript{4}{Zhejiang University}, \\
 \texttt{\{ywang26, andyhci, gxing, huiliu, hwang\}@uw.edu},\\
 \texttt{juiteh@cs.cmu.edu}, 
 \texttt{shaunkuan.10@nycu.edu.tw},\\ 
 \texttt{\{qiqian.21, chiming.21, gaoangwang\}@intl.zju.edu.cn}, \\
 \texttt{shengyuhao@zju.edu.cn} \\
 * Equal contribution
}

\begin{document}

\maketitle

\begin{abstract}
  Sensor fusion is crucial for an accurate and robust perception system on autonomous vehicles. Most existing datasets and perception solutions focus on fusing cameras and LiDAR. 
  However, the collaboration between camera and radar is significantly under-exploited. The incorporation of rich semantic information from the camera, and reliable 3D information from the radar can potentially achieve an efficient, cheap, and portable solution for 3D object perception tasks. 
  It can also be robust to different lighting or all-weather driving scenarios due to the capability of mmWave radars. In this paper, we introduce the CRUW3D dataset, including 66K synchronized and well-calibrated camera, radar, and LiDAR frames in various driving scenarios. 
  Unlike other large-scale autonomous driving datasets, our radar data is in the format of radio frequency (RF) tensors that contain not only 3D location information but also spatio-temporal semantic information. 
  This kind of radar format can enable machine learning models to generate more reliable object perception results after interacting and fusing the information or features between the camera and radar. 
\end{abstract}

\section{Introduction}

% sensor introduction
Reliability and robustness are critically important to autonomous driving and advanced driver assistance systems (ADAS). To achieve reliable object perception, different sensor modalities are usually considered on autonomous vehicles \cite{levinson2011towards,yang2021fast}, e.g., camera, LiDAR, radar, etc. These sensors have their own strengths and weaknesses. As one of the primary sensors, \textit{camera}, can provide rich and human-understandable semantic information with relatively high resolution. But it is sensitive to adverse weather or different lighting conditions. \textit{LiDAR}, as a popular 3D sensor for perception tasks, can provide accurate 3D point clouds for spatial analysis. However, LiDAR is not reliable in adverse weather conditions like rain and fog, which can interfere with laser light, potentially resulting in ghost obstacles in the scene. On the other hand, millimeter wave (mmWave) \textit{radar} is a traditional automotive sensor for distance and speed estimation. But radar is usually blamed for its low resolution, over-sensitivity, and poor semantic information involved. 

% sensor fusion: camera and lidar
Therefore, sensor fusion has been considered as a solution to take advantage of these different sensors and eliminate their shortcomings. Recently, some researchers have focused on sensor fusion between the camera and LiDAR \cite{qi2018frustum,ku2018joint,pang2020clocs} by either fusing the features from the camera and LiDAR during the intermediate stage or fusing the detection results at the final decision stage. However, with LiDAR involved, the system usually becomes equipment-complex and computation-expensive. This kind of solution is neither efficient nor robust to adverse driving conditions. 

% sensor fusion: camera and radar
Radar, on the other hand, is a cost-efficient sensor that can compensate for the limitations of the camera by providing robust 3D distance and speed information, which is potentially useful for autonomous or assisted driving systems. 
Two kinds of data representations are usually considered for the mmWave radar, i.e., radio frequency (RF) tensor and radar points. 
RF tensor is a dense and informative data representation, containing both amplitude and phase information, but the location and speed are implicit. Whereas radar points are explicit representations, which are usually sparse (less than 5 points on a nearby car) \cite{nuscenes2019, feng2020deep} and non-descriptive.
Traditionally, mmWave radar is often used as a supplementary sensor due to its difficulty in parsing useful clues for semantic understanding, which limits its potential for sensor fusion with other modality data. 
Some research works focus on semantic understanding, e.g., object classification and detection, on radar data~\cite{major2019vehicle,palffy2020cnn,ouaknine2020carrada,dong2020probabilistic,wang2021rodnet2,wang2021rethinking}, and joint 3D object detection and tracking~\cite{cheng2023centerradarnet}, 
These semantic understanding tasks require radar RF tensors as the input data format since more object-level information is preserved. 
However, limited public datasets include radar RF tensors with proper annotations, as discussed in Table~\ref{tab:dataset_comparison}. 

\begin{figure}[t]
    \centering
    \includegraphics[width=\linewidth]{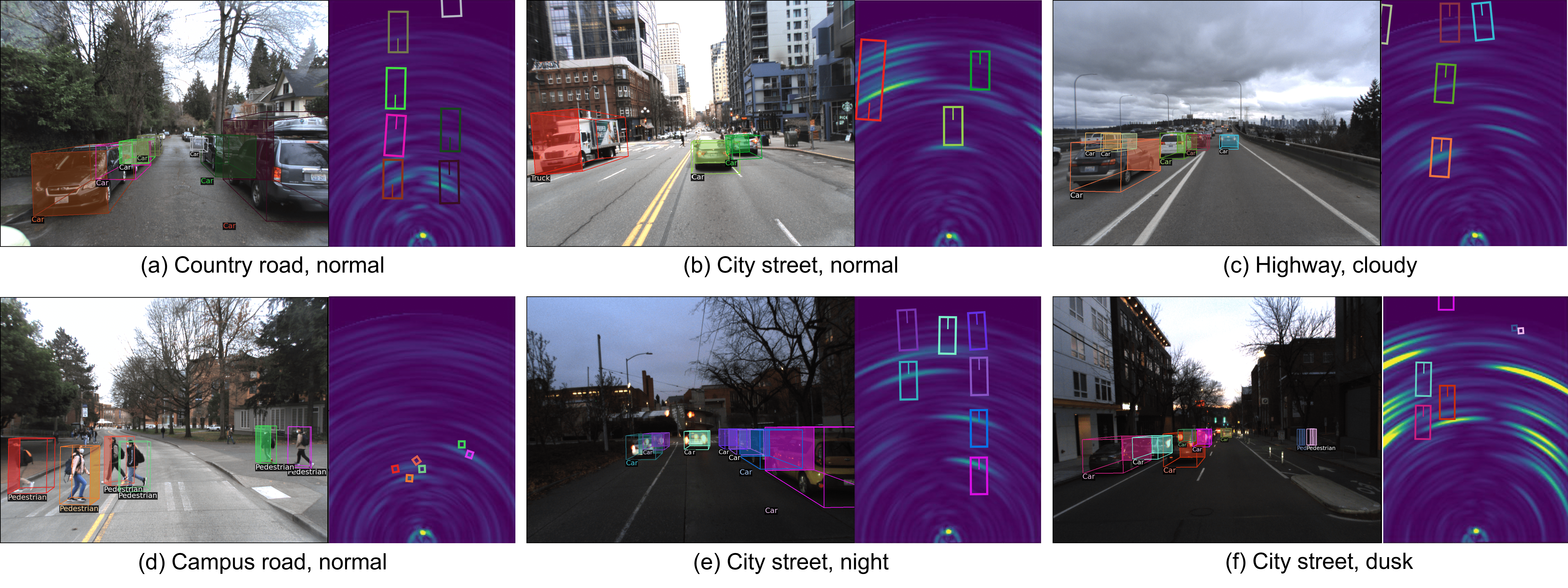}
    \caption{Examples in the CRUW3D dataset. Each example contains a camera RGB image and a radar RF tensor pair. The RF tensors are transformed to Cartesian coordinates for better visualization. We include data examples under different driving scenarios and lighting conditions. The corresponding 3D bounding box annotations are projected to RGB and RF tensors, respectively. }
    \label{fig:data_ex}
\end{figure}

To fill up the lack of such data and annotation, we introduce a new dataset, named CRUW3D, containing 66K synchronized camera, radar, and LiDAR data frames, under various driving scenarios, with object 3D bounding box and trajectory annotations. 
Fig.~\ref{fig:data_ex} shows some examples of our data and annotations in CRUW3D. 
To enhance the precision of data labeling, we include a LiDAR in our data collection system. 
Based on LiDAR point clouds, we carefully label the object 3D bounding boxes in each time frame and the object trajectories throughout the temporal sequences.
We also provide the calibration parameters among sensors to allow data/information transformation among different modalities or for sensor fusion setups. 
We hope the CRUW3D dataset will enable more research on reliable and robust collaborative perception. The CRUW3D dataset will be publicly available soon.

Overall, our CRUW3D dataset has the following key contributions:
\begin{itemize}
    \item It is the first public dataset with synchronized camera RGB images, raw radar analog-to-digital-converter (ADC) data, radar RF tensors \textit{with phase}, and LiDAR point clouds, to the best of our knowledge. 
    \item It includes object annotations of 3D bounding boxes and 3D object trajectories, which is valuable for various object perception tasks, e.g., 3D object detection and 3D multi-object tracking.
    \item It contains different lighting conditions, which are challenging for vision-based object perception methods, thus, providing a good benchmark for sensor-fusion-based object perception algorithms.
\end{itemize}

\section{Related Works}

% \subsection{Autonomous Driving Datasets}

Autonomous driving datasets have attracted great attention for deep learning based object perception methods. The KITTI dataset \cite{geiger2013vision} is the first complete autonomous driving dataset, which includes stereo cameras and a LiDAR, with various annotations. Recently, larger-scale and more advanced datasets are available, e.g., BDD100K~\cite{yu2020bdd100k}, nuScenes~\cite{nuscenes2019}, ApolloScape~\cite{huang2019apolloscape}, Waymo Open~\cite{sun2020scalability}. However, due to the hardware compatibility and less developed radar perception techniques, most datasets do not incorporate radar signals as a part of their sensor systems. 

% \subsection{Datasets with mmWave Radar}

Among the available radar datasets, some of them, e.g., nuScenes~\cite{nuscenes2019}, HiRes2019~\cite{meyer2019automotive}, RadarRobotCar~\cite{RadarRobotCarDatasetICRA2020}, RADIARE~\cite{sheeny2020radiate}, etc., consider radar data in the format of radar points that do not contain the useful Doppler and surface texture information of objects for semantic understanding. 
Other researchers focus on using RF tensors as the radar data format. 
Specifically, some manage to collect a dataset with camera, radar and LiDAR, and annotate the objects as 3D bounding boxes based on the dense point cloud from LiDAR \cite{major2019vehicle,dong2020probabilistic}. Others consider the camera-radar solution without a LiDAR \cite{ouaknine2020carrada,palffy2020cnn}, whose annotation format is usually at pixel or point level. However, most of the datasets are not publicly available, as shown in Table~\ref{tab:dataset_comparison}. 

A recent dataset, K-Radar~\cite{kradar}, provides camera images, radar RF tensors, and LiDAR point clouds with 3D bounding boxes and tracking annotations in diverse weather and lighting conditions. 
However, their radar RF tensors only contain magnitude response. 
Compared with their radar data format, our radar data contains not only amplitude values but also phase response, which provides semantics useful for classification and scene understanding. 
Moreover, we provide radar data in its raw format, called analog-to-digital converter (ADC). 
Radar's ADC data inherently keeps more information.
We foresee a more powerful architecture to consume radar ADC data and bypass the need for time-domain to frequency-domain conversion and, as a result, achieve better 3D perception capability.

\begin{table}
  \caption{Comparison with related datasets with radar data by modality, data format, scenario, etc. Better settings are marked in gray. Our CRUW3D dataset can fill the gap among these related datasets with RF radar tensors, ADC samples, and 3D bounding boxes with tracking identities.}
  \label{tab:dataset_comparison}
  \centering
  \small
  \begin{tabular}{lcccccccc}
    \toprule
    Dataset & Modality\tablefootnote{Modalities: ``C'' for camera, ``R'' for radar, ``L'' for LiDAR.} & Radar\tablefootnote{Radar data formats: ``RP'' for radar points, ``RF'' for radio frequency (RF) tensors.} & Scenario & Scale & Class & Anno & Public \\
    \midrule
    nuScenes \cite{nuscenes2019} & \cellcolor{Gray} C/R/L & RP & \cellcolor{Gray} combined & 5.5 hours & 23 & 3D Box+Trk & \cmark \\
    RADIATE \cite{sheeny2020radiate} & \cellcolor{Gray} C/R/L & RP & \cellcolor{Gray} combined & 3 hours & 7 & 2D Box & \cmark \\
    HiRes2019 \cite{meyer2019automotive} & \cellcolor{Gray} C/R/L & RP & normal & 546 frames & 7 & 3D Box & \cmark \\
    CARRADA \cite{ouaknine2020carrada} & C/R &  RF & normal & 21.2 min & 3 & Pixel & \cmark \\
    RTCnet \cite{palffy2020cnn} & C/R &  RF & normal & 1 hour & 3 & Point & \cellcolor{Red} \xmark \\
    RADDet \cite{zhang2021raddet} & C/R &  RF & normal & 10K frames & 6 & 2D Box & \cmark \\
    CRUW \cite{wang2021rethinking} & C/R &  RF & \cellcolor{Gray} combined & 3.5 hours & 3 & Point & \cmark \\
    ROD2021 \cite{wang2021rod2021} & C/R &  RF & \cellcolor{Gray} combined & 28 min & 3 & Point & \cmark \\
    Qualcomm \cite{major2019vehicle} & \cellcolor{Gray} C/R/L & RF & normal & 3 hours & 1 & 3D Box & \cellcolor{Red} \xmark \\
    Xsense.ai \cite{dong2020probabilistic} & \cellcolor{Gray} C/R/L &  RF & normal & 34.2 min & 1 & 3D Box & \cellcolor{Red} \xmark \\ 
    K-Radar \cite{kradar} & \cellcolor{Gray} C/R/L & RF & \cellcolor
{Gray} combined & 58.3 min & 5 & 3D Box+Trk & \cmark \\ 
    \midrule
    \textbf{CRUW3D} (Ours) & \cellcolor{Blue} C/R/L & \cellcolor{Blue} RF\&ADC & \cellcolor{Blue} combined & 40 min & 5 & 3D Box+Trk & \cmark \\
    \bottomrule
  \end{tabular}
\end{table}

\section{CRUW3D Dataset}

% In this section, we introduce our CRUW3D dataset including dataset collection hardware and software systems, data processing method, data annotation procedure, and dataset statistics.  

\subsection{Data Collection}

We propose a dataset collection pipeline with stereo cameras, mmWave radar, and a LiDAR, including a sensor platform, a data collection software, and a sensor calibration method. With our proposed pipeline, the data collected from three sensor modalities can be temporally synchronized and spatially calibrated accurately. 

\begin{figure}
  \centering
  \includegraphics[width=.9\linewidth]{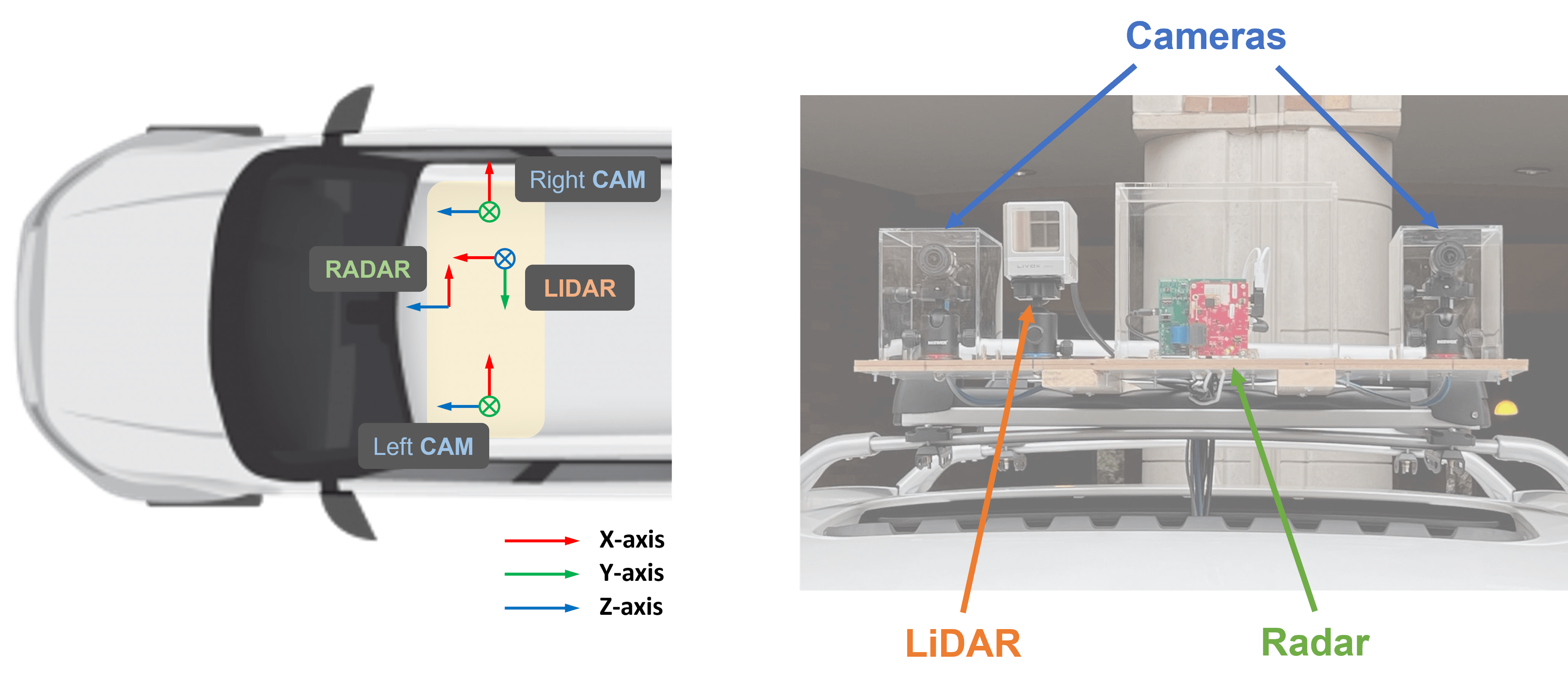}
  \caption{Sensor coordinates and sensor platform for our CRUW3D dataset, including two cameras, one mmWave radar, and one LiDAR. Note that our radar does not have elevation angle resolution (i.e., $y$-axis), thus, it is equivalent to the BEV of the camera after applying a translation vector between the two sensors.}
  \label{fig:hardware}
\end{figure}

\paragraph{Sensor Platform}
Our dataset collection sensor system is shown in Figure~\ref{fig:hardware}. There are two FLIR BFS-U3-16S2C-CS cameras, one TI AWR1843 radar board, and one Livox Horizon LiDAR. The detailed specifications are listed in Table~\ref{tab:hardware}. 

\begin{table}
    \caption{Sensor Configurations for CRUW3D Dataset.}
    \label{tab:hardware}
    \centering
    \small
    \begin{tabular}{l c | l c | l c}
        \toprule
        Cameras & Value & Radar & Value & LiDAR & Value \\
        \midrule
        Frame Rate & 30 FPS & Frame Rate & 30 FPS & Frame Rate & 10 FPS\tablefootnote{The frame rate is after the point cloud integration to scan over LiDAR's field of view (FOV). Details introduced in Section~\ref{sec:data_proc}.} \\
        Pixels (W$\times$H) & 1440$\times$1080 & Frequency & 77 GHz & Point Rate & 240,000 pts/s \\
        Resolution & 1.6 MP & \# of Transmitters & 2 & Detection Range & 260 m \\
        Field of View & 93.6$^{\circ}$ & \# of Receivers & 4 & Range Precision & 0.3 cm\\
        Stereo Baseline & 0.6 m & \# of Chirps per Frame & 255 & Field of View & 81.7$^{\circ}$ $\times$ 25.1$^{\circ}$ \\
        && Max Range & 30 m & Angular Precision & 0.05$^{\circ}$\\
        && Range Resolution & 0.23 m \\
        && Min \& Max Angle & $\pm$90$^{\circ}$\tablefootnote{Better radar performance and resolution within $\pm$60$^{\circ}$.} \\
        && Azimuth Resolution & $\sim$15$^{\circ}$ \\
        \bottomrule
    \end{tabular}
\end{table}

\paragraph{Sensor Synchronization}
Our dataset collection software is based on Robot Operating System (ROS) under Ubuntu. For cameras and LiDAR, since they provide open-source API, we directly integrate them into the ROS system. However, TI only provides software based on Windows and MATLAB. Therefore, we create a Windows virtual box in our Ubuntu system and communicate different processes through ROS. We set up hardware time synchronization between cameras and LiDAR using a Transistor-Transistor Logic (TTL) signal generated by the right camera. Both camera and LiDAR sensors support TTL signal time synchronization through their APIs. On the software level, we use the ApproximateTime synchronization policy provided by the ROS library to align three sensors' data into 30 FPS time slots. To synchronize between radar and other sensors, we use a software trigger to start a data sequence collection. A service client triggers the collection process of radar data and starts another collection process of other sensor data upon receiving a response. From our experiments, the latency of the software trigger is under a few milliseconds, which is negligible.
% \todo{ray: add time synchronization description and analysis. Add some figures/numbers to show the synchronization error if available.}
More details of our data collection system are described in the supplementary document. 

\paragraph{Sensor Calibration}
First, we calibrate the stereo cameras using Zhang's method \cite{zhang2000flexible}, which gives us the intrinsic parameters, distortion coefficients, and extrinsic parameters of the two cameras. These results are used later for stereo rectification in Section~\ref{sec:data_proc}. 
For the sensor calibration between cameras and LiDAR, we adopt the calibration algorithm proposed by Dhall et al.~\cite{dhall2017lidar}. This will give us two transformation matrices, representing the transformation between the left camera and LiDAR, and between right camera and LiDAR, respectively. As for radar, which is carefully mounted and aligned with cameras and LiDAR according to their pitch angle, its coordinates are parallel to the camera's bird's-eye view (BEV). The translation vectors between the sensors are also measured to form the full transformation matrices between cameras and radar. 

\subsection{Data Processing}\label{sec:data_proc}

\paragraph{Camera Data Processing}
The image sequences captured by the stereo cameras are first undistorted and rectified based on the camera calibration. Then, for the low-quality images due to adverse lighting conditions, we conduct image enhancement to improve the quality and lighting stability of the collected videos. Here, we implement a deep learning based method, named RRDNet \cite{zhu2020zero}, to restore the underexposed image in zero shot using a three-branch CNN. To achieve stable enhancement results for video sequences, we train the network using only the first frame of each sequence, and do inference on the rest frames.

\paragraph{Radar Data Processing}
% Our radar data processing can be divided into two step, i.e., RF images in range-azimuth (RA) coordinates, and range-azimuth-Doppler (RAD) coordinates. 
% \todo{add data Cartesian coordinate transformation}

% \paragraph{RF-RA images} 
Our radar data processing is similar to the pre-processing mentioned in \cite{wang2021rodnet}, where RF tensors in radar range-azimuth coordinates are described as a bird's-eye view (BEV) representation, where the $x$-axis denotes azimuth (angle) and the $y$-axis denotes range (distance).
From the raw radar data, we first implement a range fast Fourier transform (FFT) on the received chirp samples to estimate the range of the reflections. After that, we conduct a second angle FFT on the samples along different receiver antennas to estimate the azimuth angle of the reflections. 
% An example RF-RA image is shown in Fig.~\ref{fig:rfimage}(b). 
% After being transformed into RF images, the radar data is represented as a complex-valued 2D format (with real and imaginary channels) like an image (with RGB channels), which can thus be directly processed by the CNN architectures. 
Besides, we also transform the RF tensors into Cartesian coordinates for better alignment with the camera and more clear visualization. A more detailed description of our radar data processing is mentioned in the supplementary document.

% \paragraph{RF-RAD images}
% Besides the above RF images in the range-azimuth coordinates, to obtain the speed information from radar, we further process the raw radar data into RF-RAD images. First, same as the RF-RA pre-processing procedure, we use the range FFT to estimate the range of the reflections. Then, a Doppler FFT is implemented to estimate the speed at each range grid. Afterwards, the angle FFT is appended to estimate the azimuth angle. Now, we will get a 3D tensor representing in the RAD coordinates. The relative radial speed of the targets can be further estimated with the Doppler information in the RAD domain. 

% \todo{YZ: add figures.}

\paragraph{LiDAR Data Processing}
Livox LiDAR has a special laser scanning technology called non-repetitive horizontal scanning, which is significantly different from the repetitive linear scanning offered by most traditional LiDAR sensors. It accumulates the points captured inside the FOV to get denser point clouds within an integration time window. However, based on this technology, the point cloud from LiDAR cannot cover the whole FOV within a camera frame (i.e., $\sfrac{1}{30}$ seconds). To ensure every camera/radar frame has a corresponding LiDAR frame for annotation, we accumulate the point clouds captured in the consecutive three frames (i.e., $\sfrac{1}{10}$-second time window) as a complete frame, which means the frame rate of our LiDAR is 10 FPS, as mentioned in Table~\ref{tab:hardware}.

\subsection{Data Annotation}

\begin{figure}
% \centering
\hspace{-1em}
    \begin{minipage}{0.6\linewidth}
      \captionof{table}{Statistics of the CRUW3D dataset.}
      \label{tab:data_stats}
      \centering
      \small
      \begin{tabular}{l|ccc}
        \toprule
        Description & \multicolumn{3}{c}{Value}     \\
        \midrule
        Driving Time & \multicolumn{3}{c}{40 min} \\
        \multirow{2}{*}{Scenarios} & \multicolumn{3}{c}{70\% normal, 30\% adverse} \\
        & \multicolumn{3}{c}{city street, highway, sidewalk} \\
        \midrule
        & \texttt{Overall} & \texttt{Train} & \texttt{Test} \\
        \# of Frames & 66K & 56K & 10K \\
        \# of Seqs   & 74 & 56 & 18 \\
        \# of Labeled 3D Bboxes & 80K & 57K & 23K \\
        \# of Labeled 3D Tracks & 576 & 397 & 179 \\
        \bottomrule
      \end{tabular}
    \end{minipage}
    \hspace{0.5em}
    \begin{minipage}{0.4\linewidth}
      \centering
      \includegraphics[width=\linewidth]{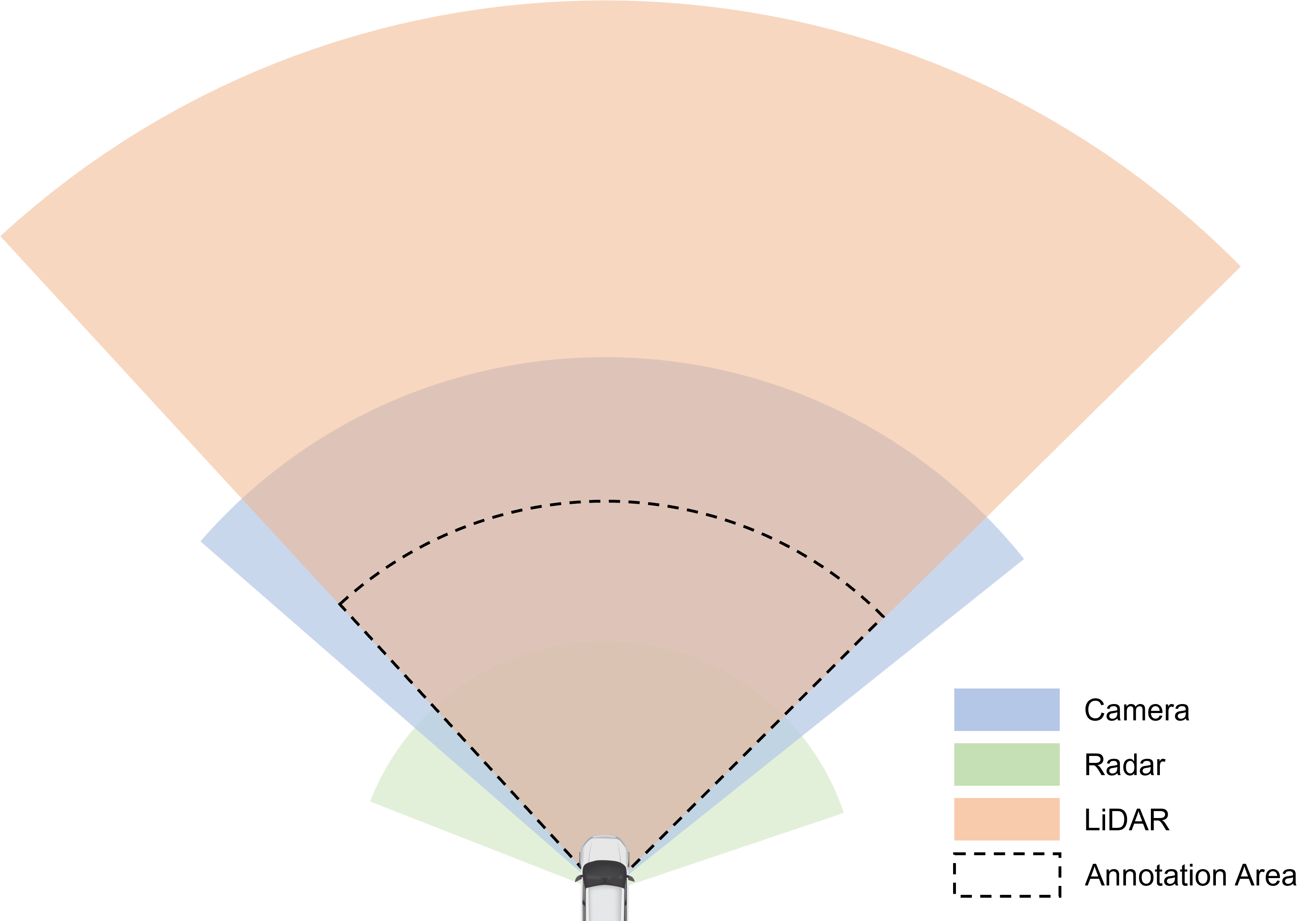}
      \captionof{figure}{Illustration of sensors' FOVs and our annotation area.}
      \label{fig:fov}
    \end{minipage}
\end{figure}

In the CRUW3D dataset, we label 3D bounding boxes on LiDAR point clouds. Unlike the 3D bounding box labels in the KITTI dataset, we use three Euler angles to represent the orientation of each bounding box, since the streets in the CRUW3D dataset are not as flat as those in the KITTI dataset. Here, we consider the following 5 object categories during the annotation: pedestrian, car, van, truck, and bus. The detailed statistics are shown in Section~\ref{sec:data_stats}.
In addition to the 3D bounding boxes, we also annotate object track IDs for later multi-object tracking (MOT) tasks. However, because different sensors have different FOVs and point clouds for the faraway objects are usually sparse, we only annotated the object within the overlapped areas, shown in Figure~\ref{fig:fov}. 
After the 3D bounding boxes are labeled on point clouds, we project all the bounding boxes to camera and radar coordinates by the transformation matrices from sensor calibration. Then, the annotations can be used to train networks for camera and radar, respectively.

\subsection{Data Statistics}\label{sec:data_stats}

Our CRUW3D dataset contains about 66K frames of synchronized camera, radar, and LiDAR data under various driving scenarios with different lighting conditions. Approximately 70\% of the data are captured in normal driving scenarios with good lighting conditions. The rest 30\% are captured in adverse lighting conditions, e.g., nighttime, or strong lighting. Some data statistics are shown in Table~\ref{tab:data_stats}. Among all the data frames, we annotate 19K frames in the training set, and 10K frames in the testing set. We use this setting for all the experiments mentioned in Section~\ref{sec:exp}.

\begin{figure}
  \centering
  \includegraphics[width=\linewidth]{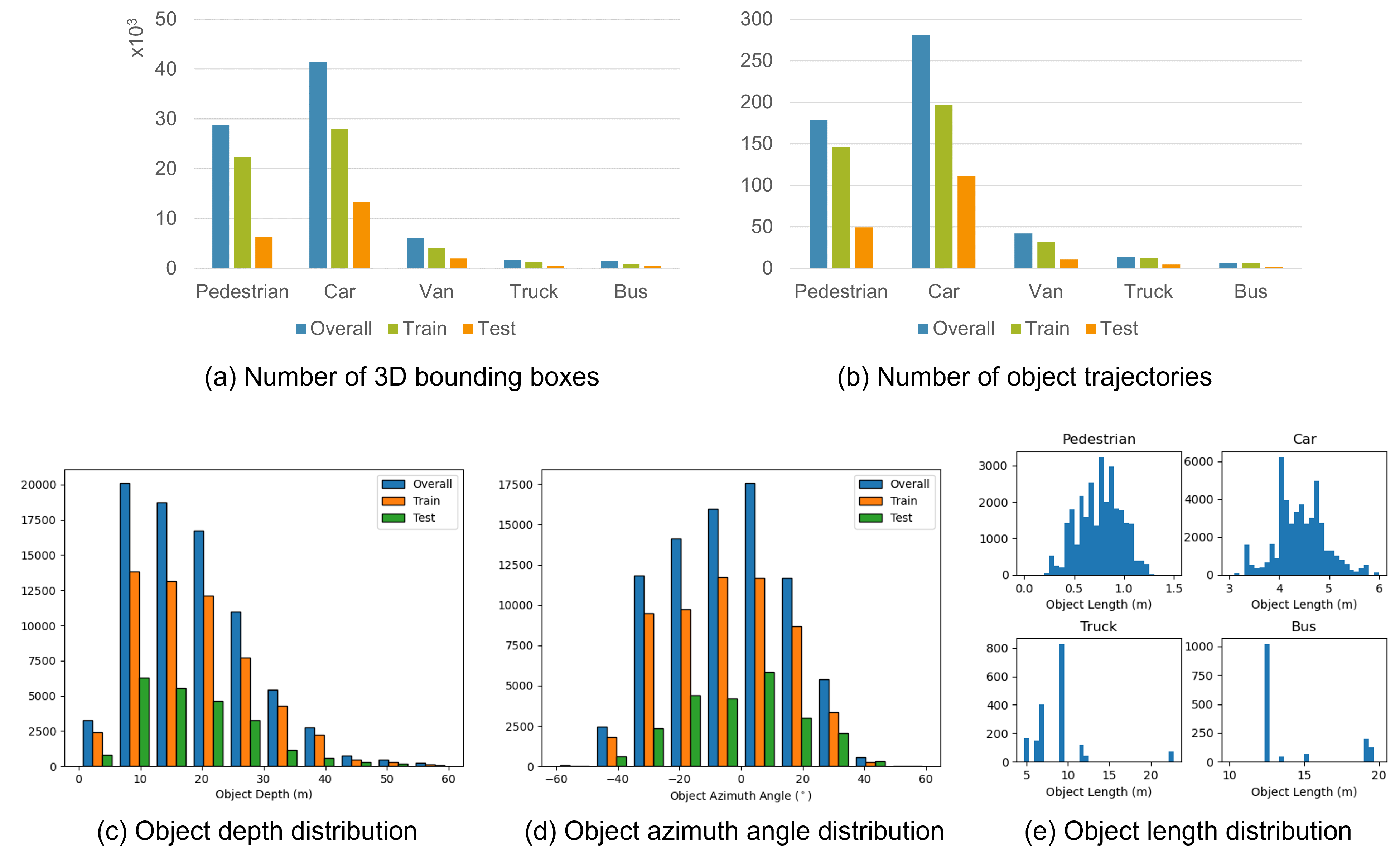}
  \caption{Object annotation distributions in the CRUW3D dataset, including (a) object 3D bounding box distribution, (b) object trajectory distribution, (c) object depth distribution, (d) object azimuth angle distribution, and (e) object length distribution. }
  \label{fig:object_stats}
\end{figure}

As for the annotations for the CRUW3D dataset, we analyze the different distributions of our labeled objects in Figure~\ref{fig:object_stats}, including the number of 3D bounding boxes, number of 3D object trajectories, object depths, object azimuth angles, and object dimensions.

\paragraph{Object Class Distribution} For the 5 object categories (i.e., pedestrian, car, van, truck, and bus) we are interested in, pedestrian and car are two dominant categories, as shown in Figure~\ref{fig:object_stats} (a) and (b), which reasonably reflects the actual object class distribution in real driving scenarios. 

\paragraph{Object Location Distribution} First, we analyze the depth distribution of the 3D bounding boxes in Figure~\ref{fig:object_stats} (c), where object depth represents the distance between LiDAR and the center of a 3D bounding box along LiDAR's $x$-axis. Here, most annotated 3D bounding boxes are distributed within 0 -- 40 meters. Besides, we also analyze object azimuth angle distribution in Figure~\ref{fig:object_stats} (d). Most labels fall into the range between $-$50$^{\circ}$ to 50$^{\circ}$, which is the overlapped region for three sensor modalities. 

\paragraph{Object Size Distribution} Figure~\ref{fig:object_stats} (e) shows the distribution of object length for different object classes, including pedestrian, car, truck, and bus. The distributions for pedestrians and cars are relatively concentrated, while those of trucks and buses are more spread. 

\paragraph{Object Trajectory} We also record some statistics for that in Table\ref{tab:data_stats}. Overall, there are 576 object trajectories, including 397 trajectories in the training set and 179 in the testing set. The average length of the object trajectories is 121 frames. 

\subsection{Comparison with Related Datasets}

We compare the CRUW3D dataset with some related datasets with radar sensors in Table~\ref{tab:dataset_comparison}. We discuss the dataset in different aspects, including sensor modalities, radar data format, driving scenarios, dataset scale, annotated object categories, annotation format, and public availability. 

From Table~\ref{tab:dataset_comparison}, most related datasets, whose radar data format is RF tensor, do not provide 3D bounding box and trajectory annotations. Although the scale of our dataset is relatively small, we validate its scale with some baseline algorithms in the experimental results in Section~\ref{sec:exp}. Nonetheless, we will continually collect and annotate data to expand the scale.

\section{Baseline Experiments}\label{sec:exp}

In this section, we conduct a series of baseline experiments on our CRUW3D dataset, including camera-based 3D object detection, camera-based 3D object tracking, radar-based object detection, and a camera-radar fusion baseline. In the following experiments, we only consider pedestrian and car as our perception target classes. % Some other experimental results are also included in the supplementary document.

% \subsection{Camera-Based 2D Object Perception}

\subsection{Camera-Based 3D Object Detection}
Monocular 3D object detection is pivotal for autonomous driving applications. 
Neural networks for 3D object detection extract images' features and detect objects in the perspective view or BEV. 
We implement SMOKE~\cite{liu2020smoke} and DD3D~\cite{park2021pseudo} as baselines on our benchmark. 

SMOKE is a single-stage 3D object detection method based on CenterNet~\cite{duan2019centernet}. Given an input image, it detects targeted objects' 3D centers projected on the image plane. However, this algorithm was originally designed for the KITTI dataset, whose 3D bounding box orientation only includes a yaw angle. We convert our quaternion-based orientation label for each bounding box to the yaw angle by ignoring the pitch and row, which assume the other rotation angles are negligible. Here, we use DLA-34 \cite{yu2018deep} as the backbone network for SMOKE during the implementation.

DD3D is built on top of another 2D object detector, named FCOS~\cite{fcos}. It uses a large-scale depth dataset DDAD15M to pre-train the network to obtain better depth-aware features from images, which achieves state-of-the-art among monocular 3D object detection methods. Here, we try two different backbone networks, i.e., DLA-34~\cite{yu2018deep} and V2-99~\cite{lee2020centermask}, during the implementation.

Similar to KITTI, the evaluation metrics include the average precision (AP) by the 3D bounding box and by BEV 2D bounding box using IOU thresholds of 0.5 or 0.7 for cars and 0.3 or 0.5 for pedestrians. The quantitative results are shown in Table~\ref{tab:exp_cam_3ddet}. From the experiments, compared with SMOKE, DD3D achieves better performance in all aspects. With the larger backbone V2-99, DD3D obtains the best performance on both cars and pedestrians.

\begin{table}
    \caption{Monocular 3D object detection baseline results on the CRUW3D testing set.}
    \label{tab:exp_cam_3ddet}
    \centering
    \small
    \begin{tabular}{l | cc|cc | cc|cc}
        \toprule
        \multirow{3}{*}{Method} & \multicolumn{4}{c|}{Car} & \multicolumn{4}{c}{Pedestrian} \\
        & \multicolumn{2}{c|}{\cellcolor{Gray} IOU=0.5} & \multicolumn{2}{c|}{\cellcolor{Gray} IOU=0.7} & \multicolumn{2}{c|}{\cellcolor{Gray} IOU=0.3} &  \multicolumn{2}{c}{\cellcolor{Gray} IOU=0.5}\\
        & AP$^{\text{3D}}$  & AP$^{\text{BEV}}$ & AP$^{\text{3D}}$ & AP$^{\text{BEV}}$ & AP$^{\text{3D}}$  & AP$^{\text{BEV}}$ & AP$^{\text{3D}}$ & AP$^{\text{BEV}}$ \\
        \midrule
        SMOKE \cite{liu2020smoke} & 44.52 & 48.55 & 17.63 & 25.11 & 10.17 & 10.57 & 2.58 & 3.36  \\
        % SMOKE (w/ Quaternion) \cite{liu2020smoke} & \\
        DD3D (DLA-34) \cite{park2021pseudo} & 57.88 & \textbf{64.86} & 24.29 & 36.09 & 16.58 & 18.62 & 6.84 & 7.44\\
        DD3D (V2-99) \cite{park2021pseudo} & \textbf{58.08} & 64.68 & \textbf{25.41} & \textbf{37.86} & \textbf{18.46} & \textbf{20.57} & \textbf{8.29} & \textbf{9.25}\\
        \bottomrule
    \end{tabular}
\end{table}

\subsection{Camera-Based 3D Object Tracking}
After object 3D detection results are predicted, we further implement a 3D multi-object tracking (MOT) algorithm, called AB3DMOT \cite{Weng2020_AB3DMOT_eccvw}, to obtain object 3D bounding box trajectories. We conduct experiments based on the 3D object detection results, i.e., SMOKE and DD3D, from Table~\ref{tab:exp_cam_3ddet} and feed into the AB3DMOT framework. AB3DMOT track different object class separately and combine them in the final stage, thus, we also evaluate the 3D MOT performance of cars and pedestrians separately, as shown in Table~\ref{tab:exp_cam_3dtrk}.  

As for the evaluation metrics for 3D MOT, we adopt the metrics proposed in \cite{Weng2020_AB3DMOT_eccvw}, including scaled average multi-object tracking accuracy (sAMOTA), average multi-object tracking accuracy (AMOTA), and average multi-object tracking precision (AMOTP). From Table~\ref{tab:exp_cam_3dtrk}, the combination of ``DD3D+AB3DMOT'' achieves the best 3D MOT performance. The performance of ``SMOKE+AB3DMOT'' on pedestrian tracking is very poor, due to the poor 3D detection quality in the previous stage.

\begin{table}
    \caption{3D MOT baseline results by AB3DMOT \cite{Weng2020_AB3DMOT_eccvw} on the CRUW3D testing set.}
    \label{tab:exp_cam_3dtrk}
    \centering
    \small
    \begin{tabular}{l | ccc|ccc}
        \toprule
        \multirow{2}{*}{3D Detector} & \multicolumn{3}{c|}{Car} & \multicolumn{3}{c}{Pedestrian} \\
        \cmidrule{2-7}
        & sAMOTA & AMOTA & AMOTP & sAMOTA & AMOTA & AMOTP \\
        \midrule
        SMOKE \cite{liu2020smoke} & 54.54 & 17.30 & 24.94 & 0.00 & -2.4 & 2.05 \\
        % SMOKE (w/ Quaternion) \cite{liu2020smoke} & \\
        DD3D (DLA-34) \cite{park2021pseudo} & 74.73 & 30.39 & \textbf{58.58} & 11.95 & 1.06 & \textbf{17.58} \\
        DD3D (V2-99) \cite{park2021pseudo} & \textbf{74.80} & \textbf{32.73} & 57.06 & \textbf{14.87} & \textbf{1.83} & 15.10\\
        \bottomrule
    \end{tabular}
\end{table}

\subsection{Radar-Based Object Detection}

\begin{wraptable}{r}{.5\linewidth}
    \vspace{-12pt}
    \caption{Radar object detection baseline results by RODNet \cite{wang2021rodnet2} on the CRUW3D testing set.}
    \label{tab:exp_rad_3ddet}
    \centering
    \small
    \begin{tabular}{l | cc}
        \toprule
        Method &  AP & AR\\
        \midrule
        RODNet (vanilla) & 28.69 & 42.37\\
        RODNet (HG) & 29.97 & 43.24 \\
        RODNet (HGwI) & 31.30 & 44.72 \\
        RODNet (HGwI + TDC) & \textbf{32.72} & \textbf{47.22} \\
        \bottomrule
    \end{tabular}
    \vspace{-5pt}
\end{wraptable}

For radar-based object detection, which detects each object as a point in the RF tensor, we use RODNet \cite{wang2021rodnet2} as our baseline method. The evaluation metrics are average precision (AP) and average recall (AR) with different object location similarity (OLS) thresholds, which are the same as our previous CRUW dataset \cite{wang2021rethinking}. 
The quantitative results are shown in Table~\ref{tab:exp_rad_3ddet}. The overall performance is lower than that on the CRUW dataset \cite{wang2021rethinking}, showing our CRUW3D dataset is much more challenging. Similar to the performance in \cite{wang2021rodnet2}, RODNet with HGwI backbone and temporal deformable convolution achieves the best performance.

\section{Conclusion}

In this paper, we introduced a new benchmark dataset, named CRUW3D, which contains synchronized and well-calibrated camera, radar, and LiDAR data with object 3D bounding box and trajectory annotations. To the best of our knowledge, it is the first public dataset with radar RF tensors with magnitude and phase information for 3D object detection and multi-object tracking tasks. With the CRUW3D dataset, sensor fusion between the camera and mmWave radar can be further exploited to improve reliability and robustness for autonomous driving.

\section{Limitations and Future Works}\label{sec:discussion}

% \paragraph{}
Although the CRUW3D dataset would contribute a lot to the sensor fusion community, dataset scale is still one of the limitations, compared with other large-scale autonomous driving datasets. Therefore, we are actively collecting and annotating more data to enlarge the scale of our dataset. 
With both labeled and unlabeled data, research works on camera and radar-based perception under the setting of self/semi-supervised learning can be further conducted.

\begin{ack}
This work is supported by CISCO Systems, Inc. [FA206070/A175367]. The authors would also like to thank the colleagues and students in the Information Processing Lab at UWECE for their help and assistance in the dataset collection, processing, and annotation works. 
\end{ack}

{
\small
\bibliographystyle{plain}
\bibliography{ref}
}

\newpage

%%%%%%%%%%%%%%%%%%%%%%%%%%%%%%%%%%%%%%%%%%%%%%%%%%%%%%%%%%%%

\appendix

\section*{Appendices}
\section{Data Collection Pipeline}

\begin{figure}[h]
    \centering
    \includegraphics[width=\linewidth]{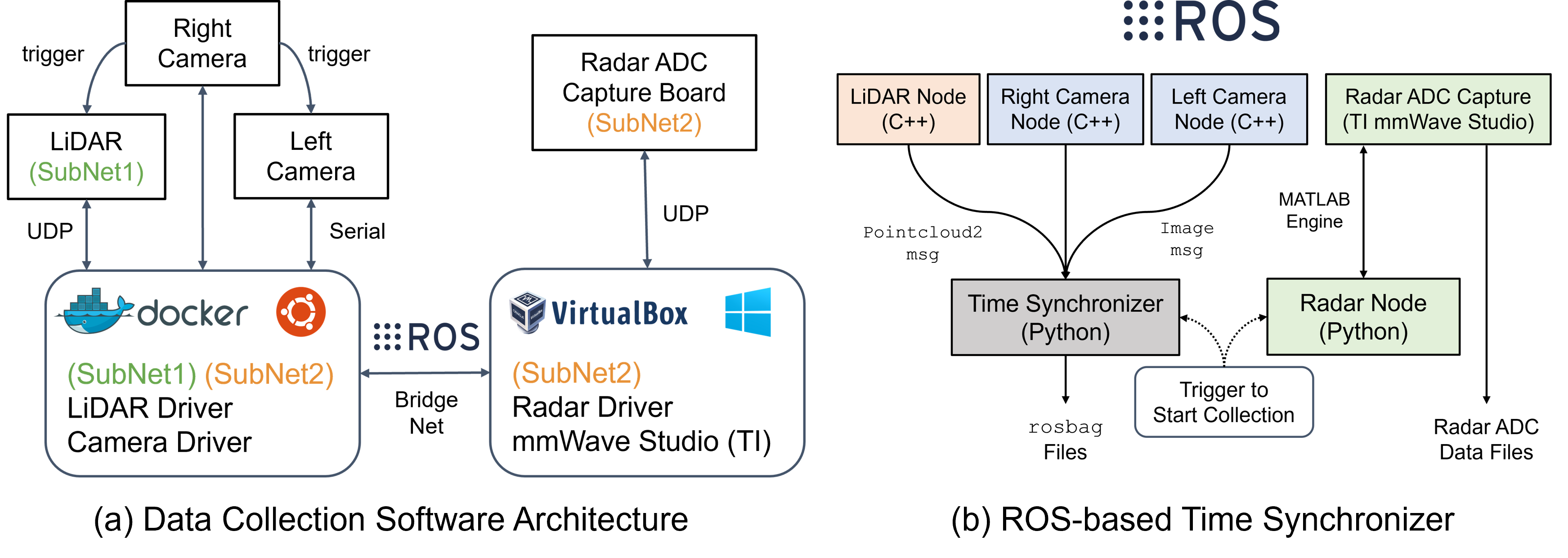}
    \caption{Pipelines for our CRUW3D dataset collection software and time synchronizer.}
    \label{fig:data_collection}
\end{figure}

The pipelines for our CRUW3D dataset collection software and the ROS-based time synchronizer are shown in Figure~\ref{fig:data_collection}. 

The main software is deployed on the Ubuntu 20.04 operating system, which is responsible for communicating with two cameras and the LiDAR. Since TI's radar does not open-source its APIs, we create a Windows 10 virtual box on the Ubuntu system to communicate with the radar. The bridge between these two systems is built by the ROS. 

As for the time synchronizer, we build the pipeline based on ROS. We use a software trigger to start time synchronizer and radar node at the same time. Then, the time synchronizer triggers the right camera, and the left camera and LiDAR are triggered through the synchronization cable (hardware trigger). On the other hand, the radar node triggers radar ADC capturing through the MATLAB engine by TI mmWave Studio.

\section{Radar Data Representations and Processing}

\subsection{Radar Data Representations}

\begin{figure}[h]
    \centering
    \includegraphics[width=.8\linewidth]{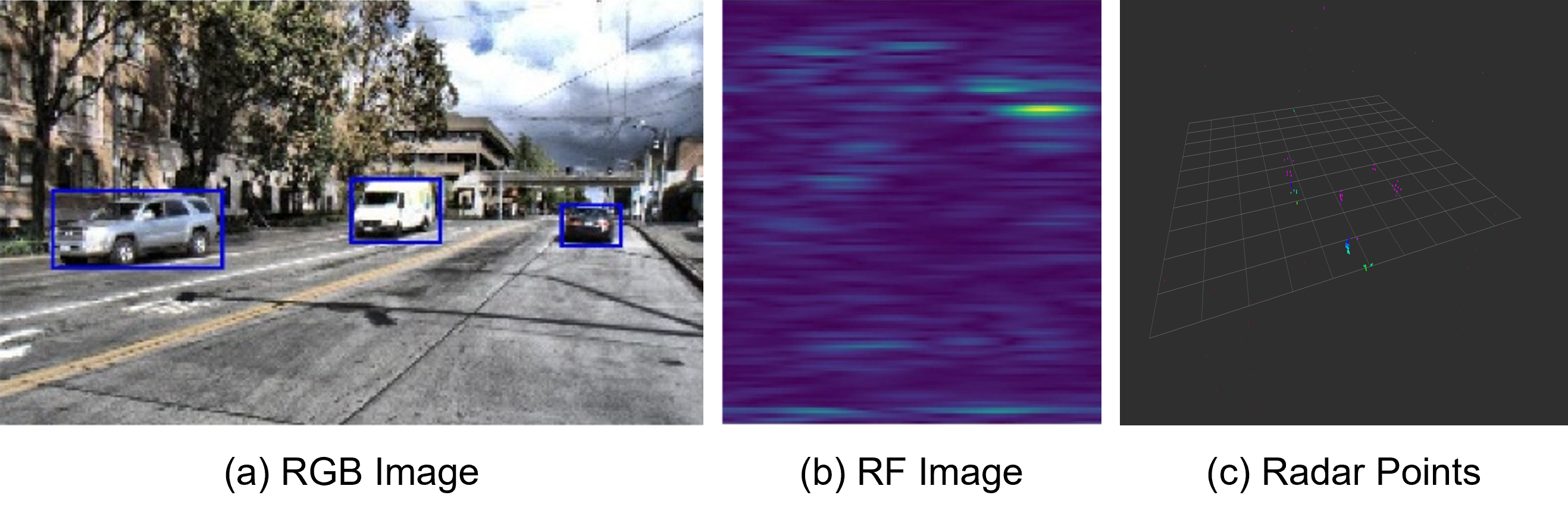}
    \caption{Visualization among (a) camera RGB image, (b) radar RF tensor, and (c) radar points. Compared with the RGB image, the semantic information in the RF tensor is implicit and hard to extract. Radar points are usually very sparse and do not contain semantic information.}
    \label{fig:data_repre}
\end{figure}

There are usually two different kinds of radar data representations, i.e., \textbf{radar points} and \textbf{radio frequency (RF) tensors}, as shown in Figure~\ref{fig:data_repre}. Radar points are more frequently used for obstacle ranging and speed estimation in autonomous driving since the ranging and speed can be directly inferred from the raw radar data through Fast Fourier Transform (FFT) and adaptive peak thresholding and clustering~\cite{richards2005fundamentals}. However, radar points from an mmWave radar are usually very sparse with relatively low angle resolution, especially compared with LiDAR \cite{caesar2020nuscenes,feng2020deep}. Thus, a large amount of useful semantic information is missing using this kind of representation. 

Radar is feasible for semantic understanding, e.g., object classification, detection, and tracking, owing to the hidden \textit{\textbf{phase}} information inside the radio frequencies. Typically, radar's signal amplitude is commonly used to estimate the distance and speed of the obstacles, while the phase information is usually not well-utilized because of its ``non-intuitiveness'', making it difficult to interpret by the classical signal processing mechanisms. 

\subsection{Radar Data Processing}

We illustrate the details of our radar data processing steps, which can be divided into two parts, i.e., RF tensors in range-azimuth (RA) coordinates to localize and classify the objects in the BEV, and range-azimuth-Doppler (RAD) coordinates to obtain the relative radial speed information.

\begin{figure}[h]
    \centering
    \includegraphics[width=.95\linewidth]{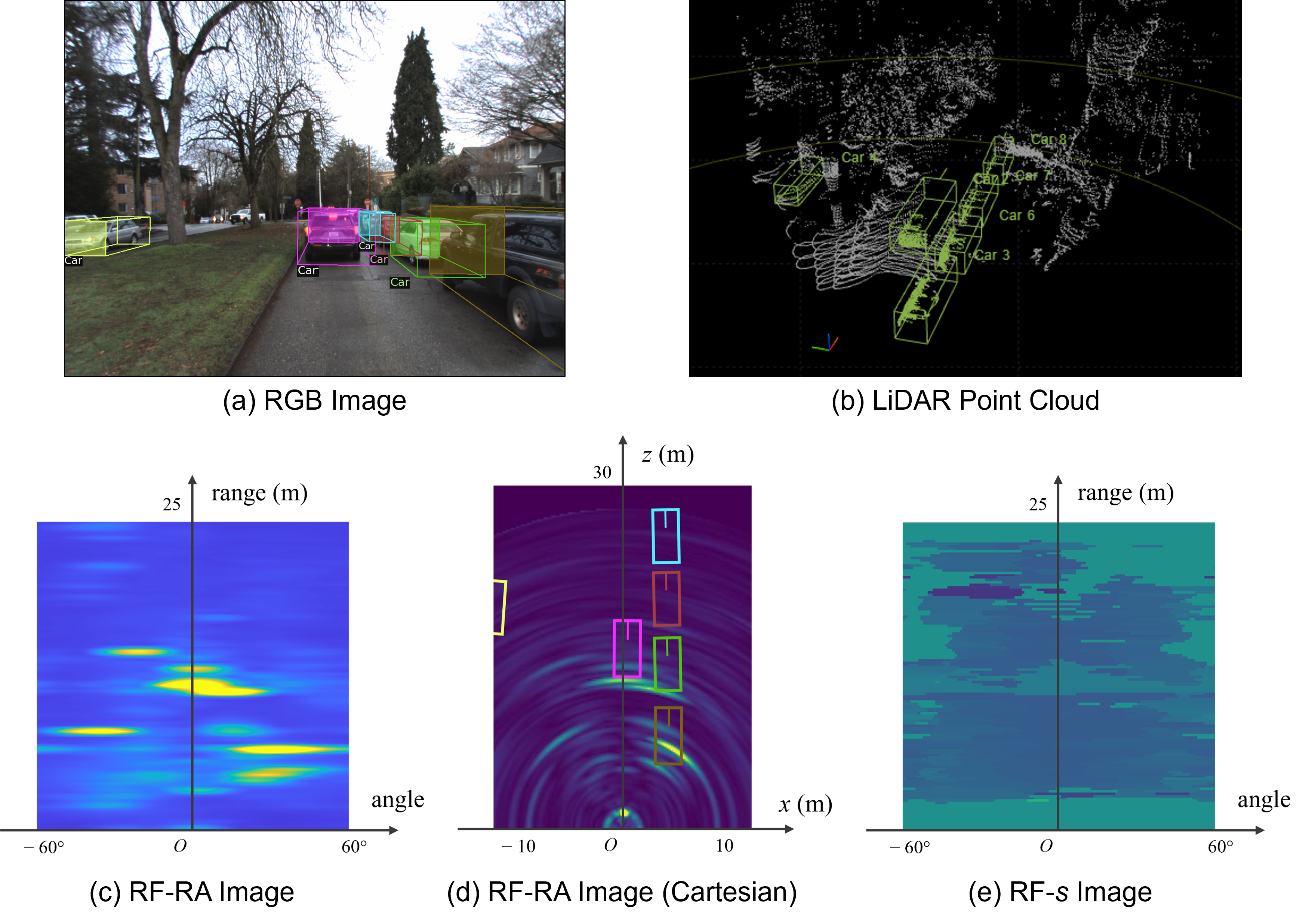}
    \caption{Visualization for multi-modality data processing, including RGB image, LiDAR point cloud, RF-RA tensor, RF-RA tensor in Cartesian coordinates, and RF-$s$ tensor for relative radial speed map. }
    \label{fig:rfimage}
\end{figure}

\paragraph{RF-RA tensors} 
This process is the same as the pre-processing mentioned in \cite{wang2021rodnet}. RF tensors in radar range-azimuth coordinates can be described as a bird's-eye view (BEV) representation, where the $x$-axis denotes azimuth (angle) and the $y$-axis denotes range (distance).
From the raw radar data, we first implement a range FFT on the received chirp samples to estimate the range of the reflections. After that, we conduct a second angle FFT on the samples along different receiver antennas to estimate the azimuth angle of the reflections. 
An example RF-RA tensor is shown in Figure~\ref{fig:rfimage}~(c). 
After being transformed into RF tensors, the radar data is represented as a complex-valued 2D format (with real and imaginary channels). 

\paragraph{RF-RA tensors in Cartesian coordinates}
To better associate the data among different sensor modalities, we also transform RF-RA tensors into Cartesian coordinates, as shown in Figure~\ref{fig:rfimage}~(d). We first generate a Cartesian grid for the target RF tensor and map each grid location to the polar coordinates. The value at a certain location is obtained by bilinear interpolation from the original RF-RA tensors. 
Note that, due to the continuity of amplitude and phase in RF-RA tensors, we conduct the interpolation on the amplitude and phase parts for each complex pixel value instead of using real and imaginary interpolation.

\paragraph{RF-RAD tensors}
Besides the above RF tensors in the range-azimuth coordinates, to obtain the speed information from radar, we further process the raw radar data into RF-RAD tensors. 
First, same as the RF-RA pre-processing procedure, we use the range FFT to estimate the range of the reflections. 
Then, a Doppler FFT is implemented to estimate the speed at each range grid. Afterward, the angle FFT is appended to estimate the azimuth angle. Here, we will get a 3D tensor representing the scenario in the RAD coordinates. In order to get the relative radial speed between the object and the ego-car, we select the speed grid with the greatest amplitude value along the Doppler axis. We call this resulting tensor speed map RF-$s$, and each element in RF-$s$ represents the relative radial speed at a certain range and angle location. An example RF-$s$ image is shown in Fig.~\ref{fig:rfimage}(e).

\section{Implementation Details}

\subsection{Monocular 3D Object Detection Implementation Details}

\paragraph{SMOKE Implementation}
We follow the original implementation of SMOKE with a few modifications.
First, we convert 3D bounding box annotations in LiDAR coordinates to camera 3D coordinates by sensor calibration. Here, we only predict the yaw angles of the objects following the original implementation of SMOKE \cite{liu2020smoke}. 
The statistics of the car and pedestrian sizes in our dataset are $[\bar{h}, \bar{w}, \bar{l}]^\intercal = [1.64, 1.88, 4.54]^\intercal$ and $[\bar{h}, \bar{w}, \bar{l}]^\intercal = [1.76, 0.73, 0.88]^\intercal$, respectively.
The statistics of object depths in our dataset is $[\bar{\sigma_z}, \bar{\mu_z}]^\intercal = [17.75, 8.84]^\intercal$.
We use the original image resolution and pad it to $1472 \times 1088$. 
We train the network with a batch size of 4 on one Tesla V100 for 80000 iterations. The learning rate is set at $2.5 \times 10^{-4}$ and drops at 50000 and 60000 iterations by a factor of $10$. During testing, we add 3D bounding box non-maximum suppression (NMS) to filter out some false positives.  
% During testing, we filter detection with a threshold of 0.25 and then apply NMS with 3D IOU criteria. 
% Our implementation platform is Pytorch 1.3.1, CUDA 10.0.

\paragraph{DD3D Implementation}
Our implementation of DD3D is mostly aligned with the original implementation. We use the depth pre-trained model provided by the authors of DD3D \cite{park2021pseudo} to train our 3D detectors. We use the same canonical 3D bounding box sizes statistics, as described in SMOKE. Color jitter, random flip, and resize were adopted in data augmentation during training. We train the network with a batch size of 4 on one Tesla V100 for 80000 iterations. The learning rate is set at $2.0 \times 10^{-3}$
and drops at 50000 and 60000 iterations by a factor of $10$. We also apply an NMS with 3D IOU criteria to filter false positives. 
% Our implementation platform is Pytorch 1.9.1, CUDA 11.1.

\subsection{RODNet Implementation Details}

Most implementation details are following the original RODNet paper \cite{wang2021rodnet2}. However, since the 3D bounding box annotations are labeled on the LiDAR point cloud, the object centers are not perfectly aligned with the reflection from the radar. Therefore, we consider the locations of the intersection between the ego-object directions and the object surfaces, i.e., the nearest points of the object surfaces from the ego-car, as the ground truth locations of the radar objects.

\end{document}